\documentclass[conference]{IEEEtran}
\IEEEoverridecommandlockouts
\usepackage{cite}
\usepackage{amsmath,amssymb,amsfonts}
\usepackage{algorithmic}
\usepackage{graphicx}
\usepackage{array}
\usepackage{booktabs}
\usepackage{longtable}
\usepackage{subcaption}
\usepackage{textcomp}
\usepackage{xcolor}

\def\BibTeX{{\rm B\kern-.05em{\sc i\kern-.025em b}\kern-.08em
    T\kern-.1667em\lower.7ex\hbox{E}\kern-.125emX}}
\begin{document}

\title{Integrated Scenario-based Analysis: A data-driven approach to support automated driving systems development and safety evaluation\\
}

\author{ Gibran~Ali,
Kaye~Sullivan,
Eileen~Herbers,
Vicki~Williams,
Dustin~Holley\\
Jacobo~Antona-Makoshi,
Kevin~Kefauver
\thanks{G. Ali*, K. Sullivan, E. Herbers, V. Williams, J. Antona-Makoshi, and K. Kefauver are with  Virginia Tech Transportation Institute (*email: gali@vtti.vt.edu)}
\thanks{D. Holley is with the Global Center for Automotive Performance Simulation}
}

\maketitle

\begin{abstract}
Several scenario-based frameworks exist to aid in vehicle system development and safety assurance. However, there is a need for approaches that combine different types of datasets that offer varying levels of case severity, data richness, and representativeness. This study presents an integrated scenario-based analysis approach that encompasses scenario definition, fusion, parametrization, and test case generation. For this process, ten years of fatal and non-fatal national crash data from the United States are combined with over 34 million miles of naturalistic driving data. An illustrative example scenario, ``turns at intersection'', is chosen to demonstrate this approach. First, scenario definitions are established from both record-based and continuous time series data. Second, a frequency analysis is performed to understand how often events from the same scenario occur at different severities across datasets. Third, an analysis is performed to show the key factors relevant to the scenario and the distribution of various parameters. Finally, a method to combine both types of data into representative test case scenarios is presented. These techniques improve scenario representativeness in two major ways: first, they populate an entire spectrum of cases ranging from routine events to fatal crashes; and second, they provide context-rich, multi-year data by combining large-scale national and naturalistic datasets. 
\end{abstract}


\section{Introduction}

Advanced Driver Assistance Systems (ADAS) and Automated Driving Systems (ADS) are designed to automate some or all driving tasks. These systems are expected to enhance driving convenience, comfort, and most importantly, traffic safety \cite{NHTSA_AV4}\cite{Peng_2024}\cite{NHTSA_Safety}. When developing and evaluating these systems, the focus has generally been on how they perform in avoiding collisions \cite{ISO34502}. Broader approaches for evaluation and development have also been proposed. For example, incorporating data from both routine driving and safety-critical events during the system development process may lead to further safety improvements \cite{Peng_2024}\cite{Schoner_2022}. 

To improve the development and evaluation of ADAS/ADS functions, situations of interest ranging from routine driving to rare events are analyzed with scenario-based methodologies \cite{Amersbach_2019}\cite{Baumler_2024_Categorization}. Human driver benchmarks, derived from crash statistics and naturalistic driving data across scenarios, are crucial references to inform the development and evaluation of these functions \cite{Scanlon_2023}\cite{Flannagan_2023}. While crash statistics provide insights into rare events such as crashes, naturalistic data offers a unique view of both routine and outlier driving scenarios encountered in real-world conditions. Integrating these datasets for the development and assessment of ADAS/ADS can provide a more representative sample of the driving environment, enabling the demonstration of a broader safety framework. 

Numerous data-driven scenario-based safety assurance frameworks exist \cite{Baumler_2024_Categorization}\cite{Roesener_2017}\cite{Antona-Makoshi_2019} and are being explored internationally \cite{ISO34502}\cite{ISO21448}\cite{SUNRISE}. These frameworks recognize that multiple data sources are needed and that an extensive data foundation enhances the representativeness of the scenario dataset \cite{Elrofai_2018}. However, existing frameworks lack explicit guidance on how to effectively integrate and operationalize data from different sources. They also do not explain how much of the complexity in the real-world driving environment is captured by the data. Different datasets contain varying levels of information, such as  the range of event severity, the context and detail of the data available, and the overall number of events and vehicle-miles travelled (VMT) within each dataset. Therefore, the dataset choices play an important role in determining the effectiveness of scenario-based development and safety assurance activities.

Recently, an effort has been made to categorize the current datasets and methods used for data-driven scenario generation \cite{Baumler_2024_Categorization}. The authors of this review paper stress the need for ``scenario fusion'', a process to combine the information from different sources into a scenario set \cite{Baumler_2024_Categorization}\cite{Baumler_2024_Fusion}. Scenario fusion enables the parametrization of crash data beyond what is obtainable solely from police accident reports (PAR). For instance, to generate a representative test scenario that falls within an operational design domain (ODD), police-reported crashes within the same ODD can be augmented with kinematic data, such as initial speeds and accelerations, obtained from a related ODD-defined naturalistic driving data source. However, one challenge in fusing data sources is ensuring that the selected ODD is clearly defined across all sources to guarantee the relevance of any extracted information.

Typical scenario generation techniques have largely relied on information derived from high-severity crashes, often involving fatalities. In contrast, naturalistic data encompasses a broader spectrum of driving events with time series data, comprising both lower-severity incidents and typical driving behaviors. Given that the purpose of extracting scenarios is to bolster the development and evaluation of ADAS and ADS by capturing real-world driving contexts—ranging from everyday driving routines to infrequent occurrences such as crashes—the integration of crash data and naturalistic driving data is essential. This integration ensures the creation of a comprehensive scenario-based assessment, enabling the examination of functional requirements and safety performance. 

In addition to providing examples of specific real-world cases for scenario generation, crash and naturalistic datasets also provide several important insights that are essential for vehicle system development and safety assurance. For example, it is important to understand how challenging a particular scenario can be in comparison to others. Similarly, it is also important to know how often a specific scenario occurs at various levels of severity ranging from routine driving to a fatal crash. Vehicle system developers can also benefit from knowing key parameter distributions and understanding whether certain factors such as lighting conditions, weather, or roadway properties play a disproportionate role in a specific scenario. These insights can equip design and safety teams to prioritize scenarios effectively and streamline development cycles.

\begin{figure*}[t]
    \centering
    \includegraphics[width=1.8\columnwidth,keepaspectratio]{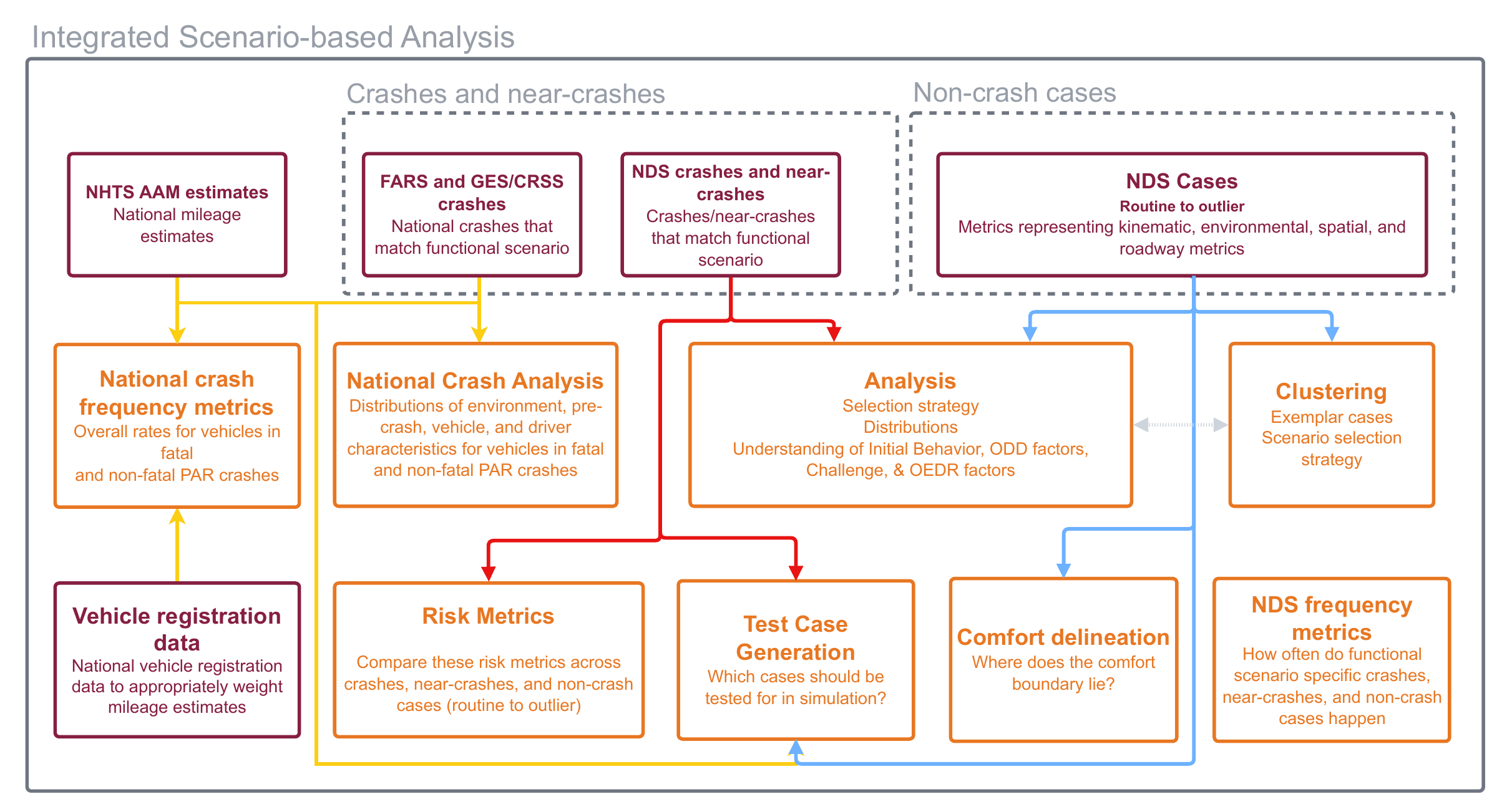}
\caption{Integrated scenario-based analysis framework consisting of datasets (maroon) and modules (orange).}
\label{fig:scenario-framework}
\end{figure*}

In this paper, we adopt data-driven scenario-based analysis \cite{Baumler_2024_Categorization}\cite{Roesener_2017}\cite{Antona-Makoshi_2019} and human benchmarking approaches \cite{Scanlon_2023}\cite{Flannagan_2023}\cite{Kusano_2023} to support the development and safety evaluation of ADAS/ADS. A framework, a set of tools, and a curated selection of data sources are provided. The data sources include 10 years of National Crash Data (NCD), over 34 million miles of naturalistic driving study (NDS) data \cite{SHRP2}\cite{Kim}, and several national- and state-level driving mileage datasets from the U.S. These sources are integrated to provide specific scenario frequencies; parameterized metrics; relevant driver, vehicle, and environmental factors; test scenarios; and human benchmark data that represent the real-world traffic environment. The data and tools provided can be used to support the development of ADAS/ADS from functional inception to performance and safety evaluation. In this paper, we demonstrate the framework and tools through an example scenario of ``turns at intersections'' and discuss the scalability of our proposed approaches to other scenarios and driving environments.

\section{Overview and Goals}

Figure \ref{fig:scenario-framework} illustrates the integrated scenario framework. The ``NDS cases'' block represents one end of the real-world driving spectrum ranging from routine events to outliers and edge cases. The ``NDS crashes and near-crashes'' block represents near-misses, minor crashes, and major non-fatal crashes. The national crashes or ``FARS and GES/CRSS crashes'' block represents the other end of the severity spectrum with non-fatal and fatal crashes. Therefore, the data sources together cover the entire range of the full spectrum of real-world driving data. Using these as a starting point, the goal of this paper is to demonstrate how this framework can answer the following questions, given a scenario is selected:
\begin{itemize}
    \item How can this particular scenario be defined for each level of severity and data source?
    \item What proportion of each data source does this scenario occupy and how frequently does it occur?
    \item What could be the major factors of interest that disproportionately affect this scenario?
    \item What is the distribution of key parameters for this scenario?
    \item How can a set of simulations be created that represent this scenario?   
    
\end{itemize}

Over the next sections, the different parts of the framework are explained and answers to the above questions are provided through illustrative examples. The various terms/acronyms used in Figure \ref{fig:scenario-framework} will be defined in the subsequent sections.

\section{Data And Methods}
Multiple datasets were required to develop data-driven methods to support scenario definition, parametrization, and scenario test case generation  that include a range of crash severities.  In this section, the crash and driving data sources will be described along with the methods used to fuse the data sources.  

\subsection{National Crash Data}

The NCD used in this analysis is based on police reports for U.S. traffic crashes. The counts of vehicles in fatal crashes are sourced from the Fatality Analysis Reporting System (FARS), and the estimates of vehicles in non-fatal crashes are sourced from the Crash Report Sampling System (CRSS) and the General Estimates System (GES). Multi-year analytic datasets were created by combining the 2010–2019 annual FARS and CRSS/GES files and accommodating the differences in the annual file structure, coded elements, and coded values.   

The 10-year crash datasets were then filtered to only include crash-involved vehicles that satisfied the following criteria. First, only light passenger vehicles (LPVs) with a vehicle model year (MY) of 1997 or later that were involved in the first harmful event of the crash were selected. Second, additional selection criteria such as excluding emergency vehicles, parked vehicles, stolen vehicles, driver absent, or vehicles pursued by the police were applied. This filtering ensured that the national crash subset was suitable for analyses pertaining to passenger vehicle development. Moreover,  resulting subset formed a cohort consistent with the Second Strategic Highway Research Program (SHRP 2) NDS, which is described in the Naturalistic Driving Studies Data section. 

\subsection{National Exposure Data} \label{sec:exposure}

An overall national mileage was needed to determine the rate of vehicle involvement in fatal and non-fatal crashes associated with a particular scenario. In the U.S., the Federal Highway Administration (FHWA) publishes the widely used national estimates of VMT \cite{Tucker}\cite{Klatko}. However, for this analysis, the national mileage estimates need to represent LPVs with MY of 1997 or newer. FHWA VMT estimates are not amenable to such a modification, and therefore an alternative method of national denominator calculation was developed from the registration-based vehicles in operation (VIO) data and average annual mileage (AAM) by vehicle age. 

\begin{equation}
VMT_{est} = \sum_{CY=a}^{b}\sum_{MY=x}^{CY+1} VIO_{(CY,MY)} \times AAM_{CY-MY} 
\label{eq:VMT}
\end{equation}

Equation \ref{eq:VMT} yields the mileage estimate ($VMT_{est}$) calculated by applying the $AAM$ to the $VIO$ by vehicle age within each calendar year (CY), and those results are summed across the calendar years (from a to b), where the vehicle age is the difference between the $CY$ and the $MY$. The $AAM$ is publicly available from the National Household Travel Survey (NHTS) and the $VIO$ is commercially available \cite{FHWA}.  

\subsection{Naturalistic Driving Studies Data}

Large-scale NDSs offer a unique context-rich perspective of driving events by capturing real-world data in a natural setting. These studies cover a wide span of event severities ranging from routine driving to severe non-fatal crashes. The SHRP 2 NDS is the largest NDS in the world, with over 3,500 participants driving approximately 34 million miles. The study collected data between 2010 and 2013 and instrumented participants’ vehicles with a data acquisition system that collected data from various installed and on-board sensors, such as cameras, radar, inertial measurement units, GPS, and vehicle Control Area Network (CAN). The data was collected from six locations across the U.S. The study collected complete trips from key on to key off and therefore comprehensively represents commuter driving behaviors \cite{hankey2016description, dingus2015naturalistic}. 

Major advantages of NDS data include the post-collection context enrichment and data processing opportunities. The SHRP 2 NDS data, housed at the Virginia Tech Transportation Institute (VTTI), was augmented by processes such as lane detection using machine vision and roadway attribute addition through map-matching. Another important data augmentation step is the discovery of crashes and near-crashes within the dataset and the creation of a baselines driving set for comparison. 

Crashes and near-crashes were identified in a two-step process. First, an algorithm flagged potential crashes by searching for signatures representing harsh accelerations, swerves, safety system activations, and so forth \cite{Perez}. Second, these flagged events were then analyzed by video reduction experts to determine whether a crash or near-crash actually occurred. For confirmed safety critical events, a detailed manual labeling process was used to code information about the event, actions of the various actors, maneuvers, driver behaviors, roadway factors, and environmental conditions. The baselines, on the other hand, were randomly selected driving epochs that went through a similar labelling processes. This ensured that behavior during crash and near-crash events could be compared with an overall representative dataset. 

Another advantage of the SHRP 2 NDS is the ability to algorithmically search for certain routine scenarios and measure exposure using the continuous time series data. Algorithms can use vehicle kinematics, GPS coordinates, roadway attributes, and radar data to find instances where specific conditions are met as well as estimate total mileage accumulated in such conditions. This allows reliable estimates of how often the scenario occurs in routine driving. 

\subsection{The Spectrum of Severity and Context}

A range of severities are represented by NCD and NDS datasets. Let's consider ``turns at intersections'', a common driving scenario, as an illustrative example that will be carried through each step of this paper. In real-world driving, the ``turns at intersection'' scenario is associated with fatal crashes, major non-fatal crashes, minor crashes, near-crashes, edge cases, outliers, and routine driving. Important information about the scenario is contained at all levels of severity. For example, crashes may represent modes of failure and routine driving events may represent comfort and acceptance criteria.

At the same time, different levels of severity are contained in different datasets, which in turn provide different levels of context. Table 1 summarizes the data sources, their size and severity range, the type of information contained, and the detection method. Each of these factors affects what information can be extracted and how it can be used for analysis and scenario generation.


\begin{table}[]
\centering
\caption{Summary of the data sources used in this paper}
\label{tab:datasource}

\begin{tabular}{@{}m{1cm}m{1.5cm}m{1.5cm}m{1.5cm}m{1.5cm}@{}}
\toprule
Dataset &
  Size &
  Severity Range &
  Data type &
  Source and Detection \\ \midrule
FARS &
  $\sim$32,000 crashes per year &
  Fatal crashes &
  Records-based only &
  Police reports, 2010–2019 \\ \midrule
CRSS/GES &
  Sample of $\sim$51,000 from $\sim$6.1M crashes per year &
  Crashes that result in fatality, injury, or property damage &
  Records-based only &
  Police reports 2010–2019 \\ \midrule
SHRP 2 Crashes &
  $\sim$2,000 crashes detected and verified &
  Major to minor crashes &
  Continuous time series and records-based &
  Algorithmic search; video verification; manual labeling \\ \midrule
SHRP 2 Near-crashes &
  $\sim$7,000 near-crashes detected and verified &
  Near-crashes that required evasive maneuvering &
  Continuous time series and records-based &
  Algorithmic search; video verification; manual labeling \\ \midrule
SHRP 2 Baselines &
  $\sim$32,000 randomly sampled epochs &
  Routine driving over 5 mph &
  Continuous time series and records-based &
  Random sample; video verification; manual labeling \\ \midrule
SHRP 2 Continuous data &
  34.5 million miles of driving &
  Routine to outlier non-crash events &
  Continuous time series data only &
  Algorithmic search; verification of subsample \\ \bottomrule
\end{tabular}%

\end{table}

\subsection{The Need for Scenario Fusion from Disparate Data Sources}

FARS and CRSS datasets consist of record-based coded variables that contain information about vehicle type, maneuvers, drivers, environmental conditions, and roadway factors, among other things. These variables are populated from police crash reports based on post-event site inspections and interviews with involved parties. Such information can be used for analyzing crashes, understanding important correlating factors, and selecting cases for simulation generation. However, these datasets lack time series information about vehicle kinematics and actor trajectories, which are needed for better insight extraction and scenario generation. 

The SHRP 2 NDS crash, near-crash, and baseline datasets consist of continuous time series variables as well as coded variables similar to those of FARS and CRSS. However, the SHRP 2 NDS coded variables have been derived from driver-perspective video and other collected data sources instead of police crash reports. The 34.5 million miles in SHRP 2 NDS can also be searched through data processing algorithms looking for specific signatures such as high yaw rates indicating turns or map-based metrics indicating driving through intersections. Epochs of interest can also be summarized to extract metrics from the time series data.

Therefore, to get a holistic perspective of a scenario, a fusion of data sources that cover the spectrum of severity and context is required. National crash datasets such as FARS and CRSS and large-scale naturalistic driving datasets such as SHRP 2 NDS are complementary to each other and cover this spectrum well. Considerable effort has been made by researchers at VTTI during this and previous projects to align the record-based NCD and NDS datasets. This alignment ensures that the datasets can be analyzed side by side and appropriate conclusions can be drawn.

\section{Analysis and Results}

This section addresses the challenges of defining meaningful scenarios, focusing on creating scenario definitions for NCD and NDS datasets, particularly using ``turns at intersections'' as an example. We discuss the algorithmic approach for scenario definition in SHRP 2 NDS data and emphasize the importance of understanding the parameter space in vehicle system development. Additionally, we explore frequency estimates and concrete test scenario generation, highlighting the significance of a data-driven approach in simulation processes. 

\subsection{Definition and Selection}

The first challenge in developing a scenario with a holistic perspective is in creating meaningful scenario definitions for each dataset (NCD and NDS). Using ``turns at intersections'' as an illustrative example, a slightly different selection criteria was designed for each dataset. There were two primary selection requirements: finding cases that happened around intersections and finding cases that involved turning.  

For FARS and CRSS, a derived junction variable was created from the relation to junction variable (RELJCT2 variable, as documented in the NHTSA coding manual) \cite{CRSS_manual_2024}. The junction variable classified intersection, intersection-related, driveway access, and driveway access-related subcategories as "Junction", while all other categories were labeled as "Not a junction." For turning behavior, a derived vehicle turning variable was created from the pre-event movement, critical precrash event, and crash type (P\_CRASH1, P\_CRASH2, and ACC\_TYPE variables documented in the NHTSA coding manual). If any of the selected vehicles were turning, the case was considered to have a turn and qualified for inclusion in this scenario. Similarly, for SHRP 2 crashes, near-crashes, and baselines, the variable “relation to junction” was used to determine junction status. The variables of pre-incident maneuver, precipitating event, and motorist 2’s or 3’s pre-incident maneuvers (the last two applicable only to crashes and near-crashes) were used to determine whether the subject vehicle or another vehicle was turning. 

\begin{figure}
    \centering
    \includegraphics[width=\columnwidth,keepaspectratio]{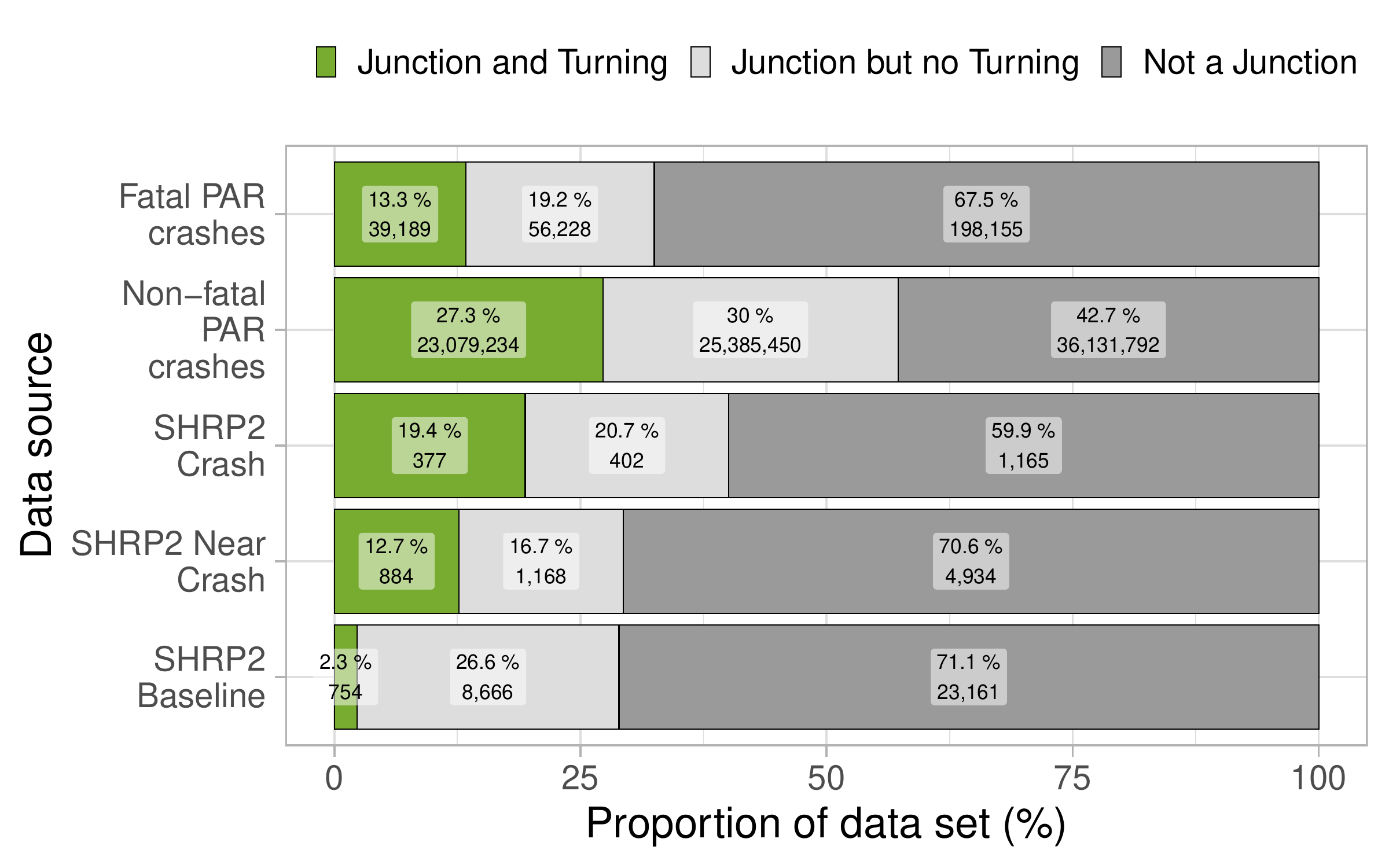}
    \caption{Proportion of each dataset that qualifies as  ``turns at intersection'' scenario.}
    \label{fig:proportion}
\end{figure}

Figure \ref{fig:proportion} illustrates the distribution of the junction and turning maneuver for selected vehicles in the 10-year fatal and non-fatal police-reported (PAR) crashes along with the SHRP 2 NDS crashes, near-crashes, and baselines.  It shows that 13.3\% of vehicles in fatal crashes, 27.3\% of vehicles in non-fatal PAR crashes, 19.4\% of SHRP 2 NDS crashes, and 12.7\% of SHRP 2 NDS near-crashes match the functional scenario definition and therefore can be used for subsequent analysis. Even though these different datasets are summarized together, it is important to understand that the subsets will not be an exact match since the data sourcing and definitions of similar variables will be different between NDS and PAR crashes. It should also be noted that this is just one way to interpret the ``turns at intersections'' scenario and different definitions may be chosen based on the exact use case of the development team.

Scenario definition and selection through algorithmic searching of all SHRP 2 NDS time series data required a different approach because of the absence of coded variables representing junction and vehicle turning maneuvers (coding was performed only on discovered crashes, near-crashes and sampled baselines). Therefore, an algorithm was developed that detected and summarized vehicle turning events. Three key metrics were selected: (1) vehicle yaw change during turn based on the integral of time series yaw rate, (2) angle between previous and next roadway segments based on underlying map geometry, and (3) vehicle yaw based on GPS heading.  Digital map metrics were used to ensure that the vehicle passed a junction with three or more roadway segments during the turn. Over 2.1 million left and right turns were detected and used in analysis.

\subsection{Parametrization, Boundary Definitions, and Outlier Detection}

Understanding the parameter space of a scenario is a fundamental step in vehicle system development. Parameter identification and definition is an iterative process that starts with assumptions of what might be relevant to the scenario and evolves through the various stages of the development cycle. Data-driven workflows have considerable potential to improve this iterative process and help developers select the right parameters and refine their definitions. 

\begin{figure}
    \centering
    \begin{subfigure}[t]{0.5\textwidth}
        \includegraphics[width=0.98\columnwidth]{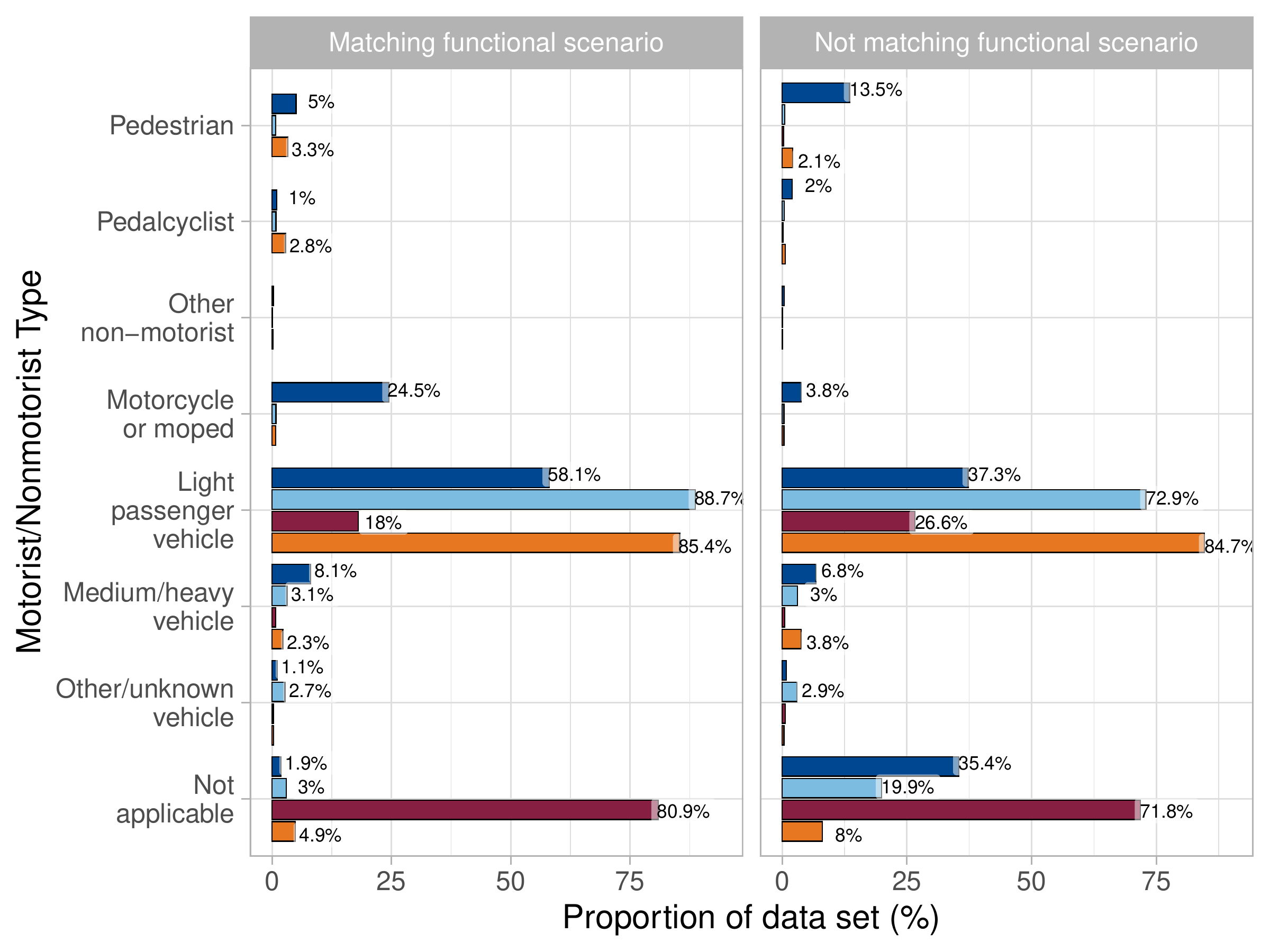}  
        \caption{Motorist/Nonmotorist Type (other involved road user)}
        \label{fig_site_selection_stats_a}
    \end{subfigure}
    \begin{subfigure}[t]{0.5\textwidth}
    \includegraphics[width=0.99\columnwidth]{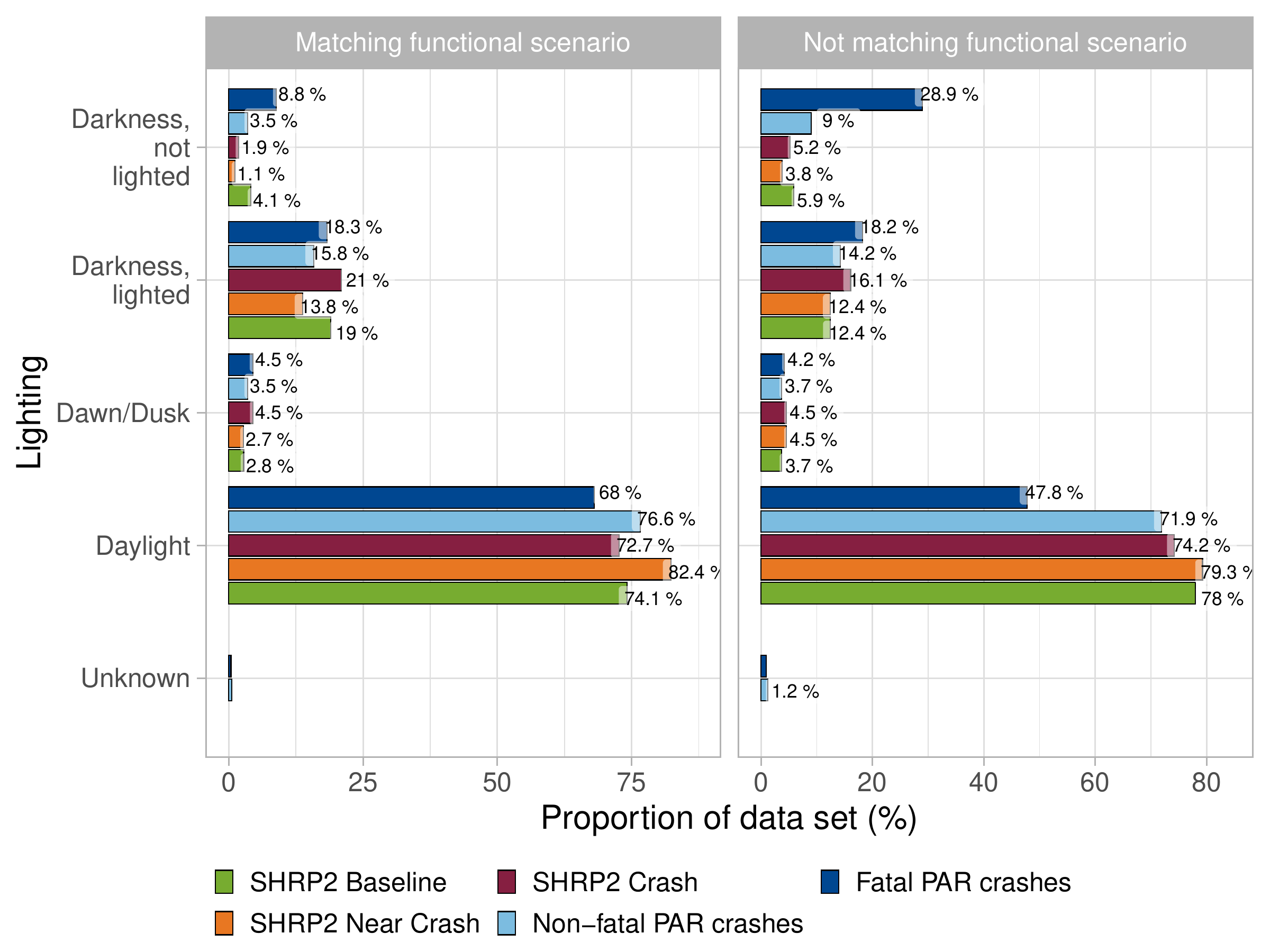}
        \caption{Lighting}
        \label{fig_site_selection_stats_b}
    \end{subfigure}
    \caption{Distribution of record-based parameters extracted from NCD and NDS data representing Motorist/nonmotorist type and lighting conditions.}
    \label{fig_site_selection_stats}
\end{figure}

Figure \ref{fig_site_selection_stats} shows the distribution of Motorist/Non-motorist Type and Lighting for cases that match the ``turns at intersection'' scenario description versus the cases that do not match the scenario description \cite{nhtsa2020}. Figure \ref{fig_site_selection_stats_a} shows that Motorcycle/moped and Medium/heavy vehicles have a much higher proportion in fatal PAR crashes than in the rest of the datasets. Similarly, SHRP 2 crashes have a much higher proportion of “Not applicable,” which are single vehicle incidents, as compared with the higher severity NCD cases. Figure \ref{fig_site_selection_stats_b} shows how lighting conditions vary across the different data sources. Such an analysis is helpful to ensure that the entire range of severity is covered and modes of failure that may be predominant at only one level of severity are not excluded. 

These plots give an example of how record-based NCD and NDS parameters can drive analysis and identify the important factors associated with a specific scenario at a particular severity level. For the NDS-NCD data fusion, the common variables allow comparisons for a variety of environmental (e.g., weather, roadway surface condition), pre-crash (e.g., trafficway flow, traffic controls), driver (e.g., impairment, distraction), and collision partner (e.g., vehicle, non-motorist, object) characteristics. 

\begin{figure}
    \centering
    \begin{subfigure}[t]{0.5\textwidth}
        \includegraphics[width=0.98\columnwidth]{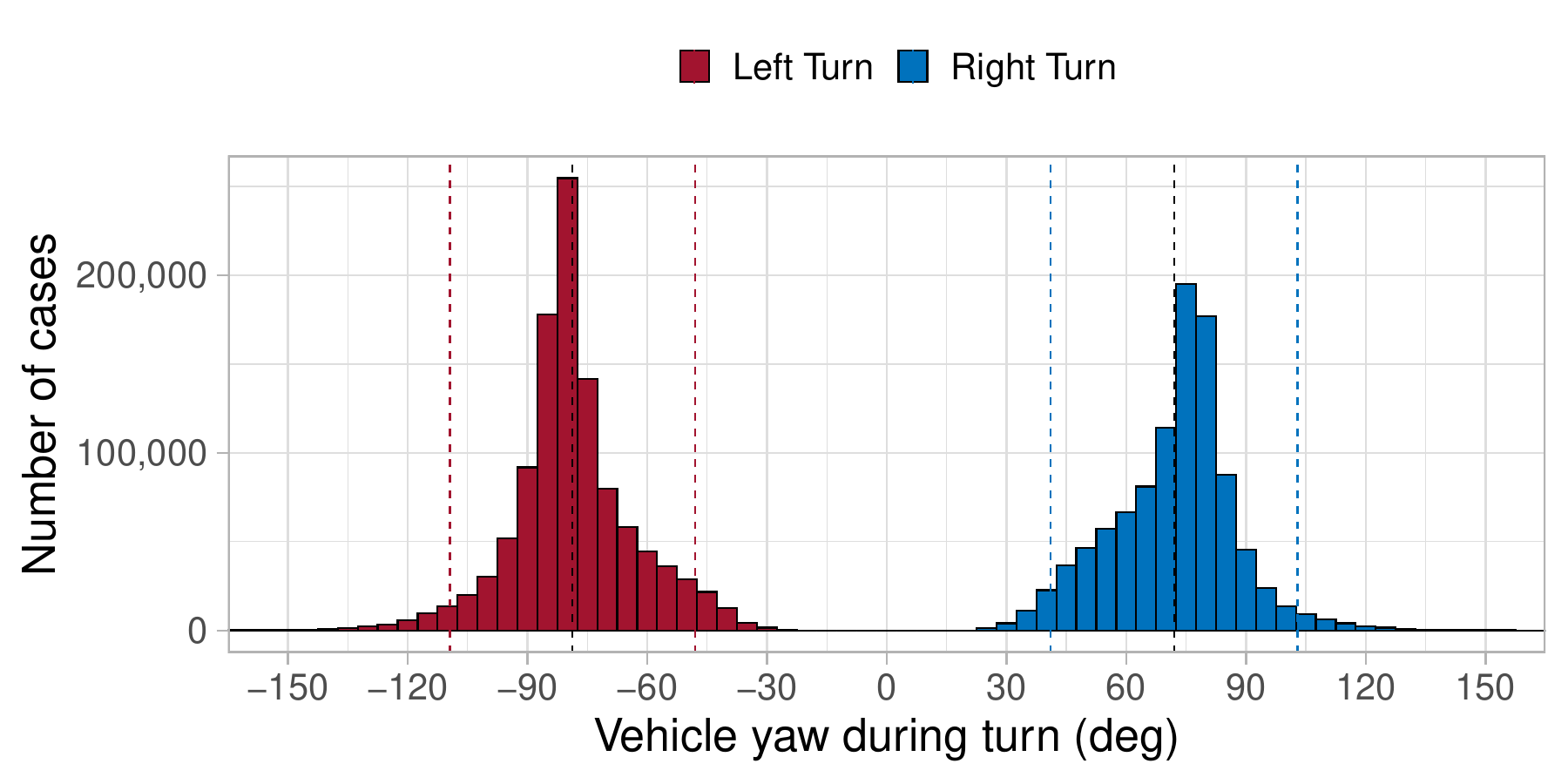}
        \caption{Histogram of net vehicle yaw during turn}
        \label{fig_yaw_accel_a}
    \end{subfigure}
    \begin{subfigure}[t]{0.5\textwidth}
    \includegraphics[width=0.98\columnwidth]{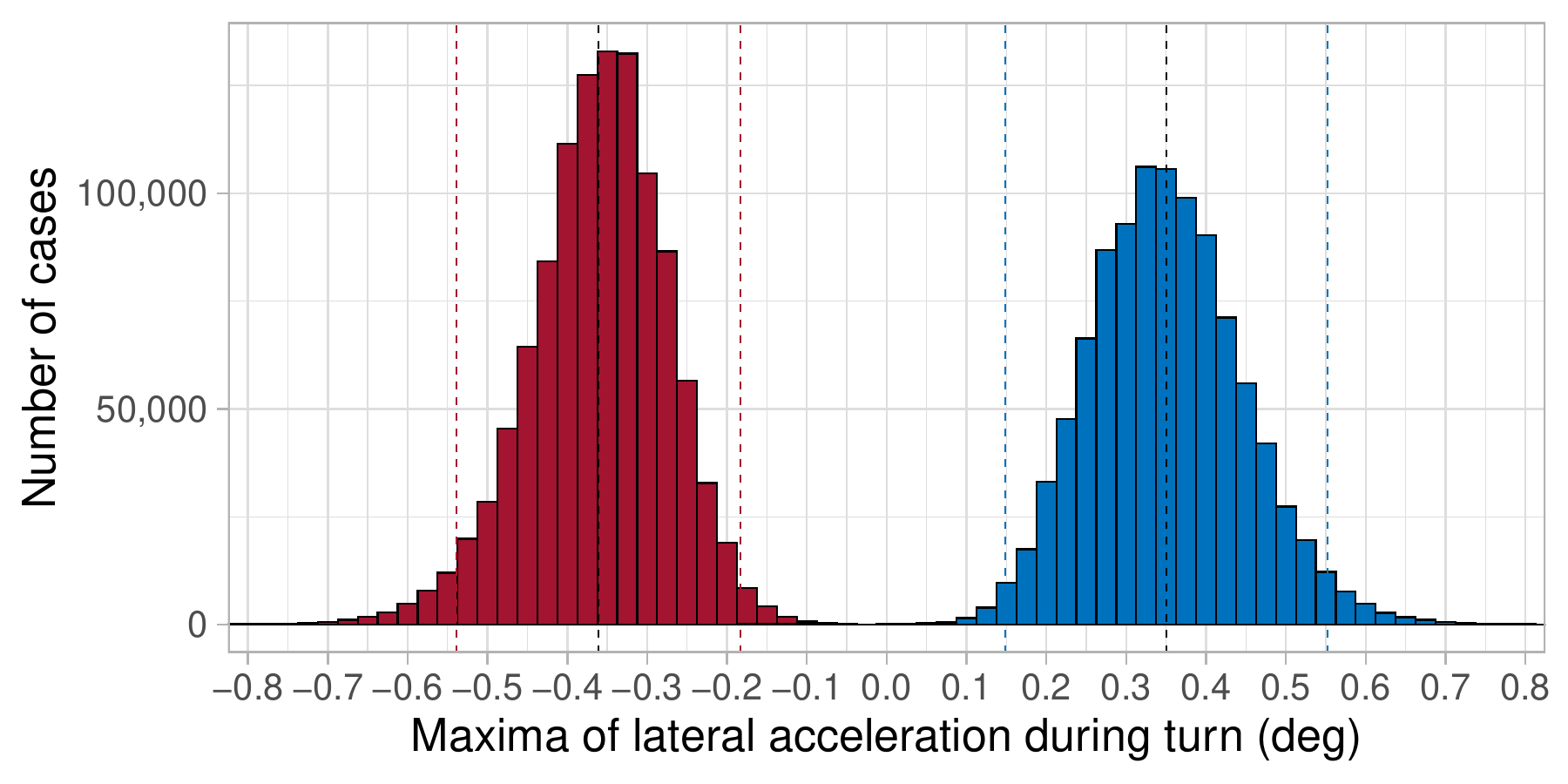}  
        \caption{Histogram of maxima lateral acceleration value during turn}
        \label{fig_yaw_accel_b}
    \end{subfigure}
    \begin{subfigure}[t]{0.5\textwidth}
        \includegraphics[width=0.98\columnwidth]{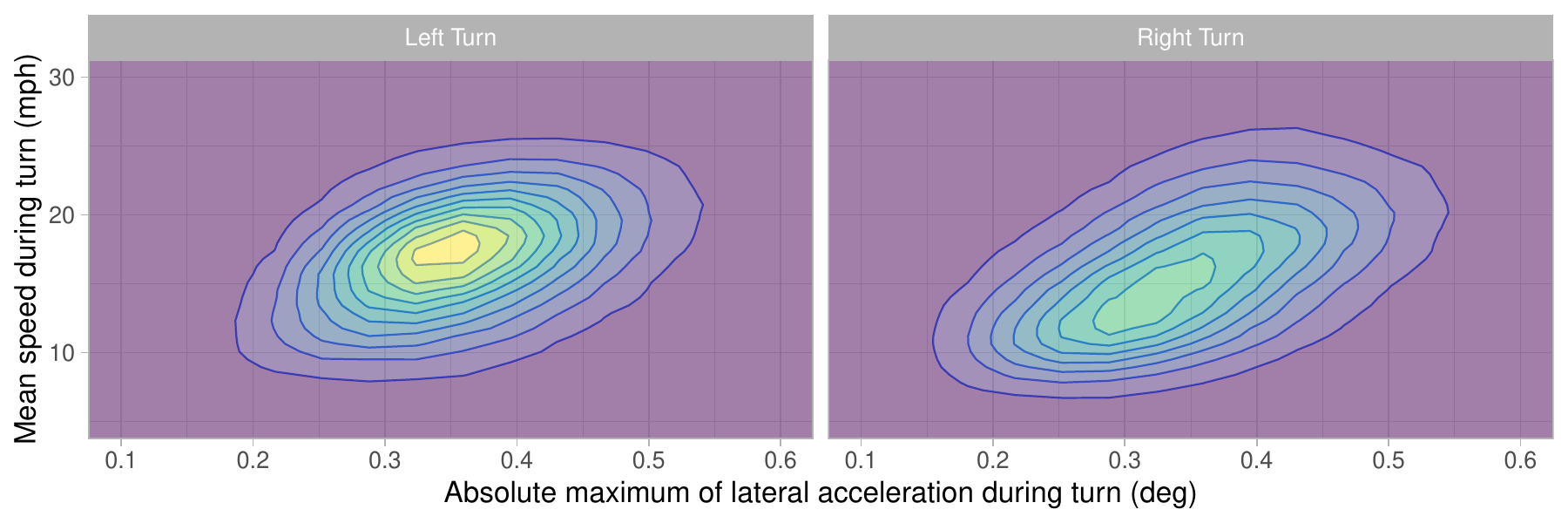}  
        \caption{Bivariate contour plot of absolute maximum lateral acceleration versus mean speed during turn.}
        \label{fig_yaw_accel_c}
    \end{subfigure}
    \caption{Important vehicle kinematic metrics describing vehicle turning behavior for over two million turns detected from SHRP 2 NDS.}
    \label{fig_yaw_accel}
\end{figure}

Figure \ref{fig_yaw_accel}a illustrates the distribution of change in vehicle yaw during the turn for over 2 million cases detected through algorithmic search. Similarly, Figure \ref{fig_yaw_accel}b shows the distribution of the maxima lateral acceleration value experienced during left and right turns. Both plots are overlaid with mean and $\pm2\sigma$ vertical lines (where $\sigma$ is the standard deviation). Parameter ranges like these can be used to set testing thresholds, develop logical scenarios, and create design specification among other things. These distributions can also be used to seed scenario development driven by coded datasets such as the NCD where such information is missing.

Figure \ref{fig_yaw_accel}c shows the bivariate distribution of absolute maximum lateral acceleration versus the mean vehicle speed during turns. Even though both left and right turns have similar peaks, it can be seen that these values are usually experienced at different speeds. During this process, the research team extracted various other metrics representing additional vehicle kinematics, driver behavior, roadway factors, environmental conditions, and other parameters of interest.  These metrics can be used to develop multivariate distributions and models, select outlier and edge cases, understand driver comfort zones, and feed into clustering processes for test case generation.

\subsection{Frequency Estimates}

Table \ref{tab:frequencyestimates} summarizes the overall, as well as the scenario specific, frequency estimates calculated from the various data sources. Care should be taken when using these measures, as they estimate different things. For NCD, the frequency estimate is based on the number of vehicles in crashes and not the number of crashes. Therefore, crashes involving two selected vehicles would be counted twice. The mileage estimates for the NCD are based on the methodology described in Section \ref{sec:exposure}. For the NDS, the frequency estimate is based on the number of crashes, and the mileage is based on the number of miles calculated from speed. Both NDS metrics are weighted to account for the differences in SHRP 2 NDS versus the national population demographics.

\begin{table}[]
\centering
\caption{Summary of estimated mileage, total number of events, number of scenario-specific events, overall rate, and scenario-specific rate for various data sources.}
\label{tab:frequencyestimates}
\begin{tabular}{@{}m{1cm}m{1cm}m{1cm}m{1.5cm}m{1.5cm}m{1cm}@{}}
\toprule
Category                & Estimated Mileage  & Total Events & Events in Scenario & Overall Rate   & Rate for Scenario \\ \midrule
LPVs in Fatal Crashes     & 25.991 Trillion miles & 293,572    & 38,856     & 1.13 per 100 MVMT & 0.15 per 100 MVMT \\
LPVs in non-fatal crashes & 25.991 Trillion miles & 84,596,476 & 23,024,145 & 325 per 100 MVMT  & 89 per 100 MVMT   \\
SHRP 2 NDS Crashes      & 36.5 Million miles & 1,720      & 321.64             & 47.12 per MVMT & 9.14 per MVMT     \\
SHRP 2 NDS Near-crashes & 36.5 Million miles & 6,982         & 768.02             & 191.3 per MVMT & 24.3 per MVMT     \\ \bottomrule
\end{tabular}%
\end{table}

The scenario-specific rates can vary by 2–3 orders of magnitude between LPVs in fatal versus non-fatal crashes. The SHRP 2 NDS’s crash rate estimate is one order of magnitude higher than the rate estimate of LPVs in non-fatal crashes. Such information helps system developers as well as safety experts to understand the severity versus frequency aspects of a scenario. These rates also provide useful human benchmarks for comparisons with ADAS/ADS systems.

\subsection{Concrete Test Scenario Generation}

Appropriately representing a scenario through parametrization is a major challenge of the simulation process in the vehicle system development cycle. Choosing an inadequate level of complexity during scenario representation can lead to insufficient feature testing, as some of the key behaviors needed to navigate the scenario may be missed. Incorporating a data-driven approach to scenario definition and parametrization can help address these issues.

\begin{figure}
    \centering
    \includegraphics[width=\columnwidth,keepaspectratio]{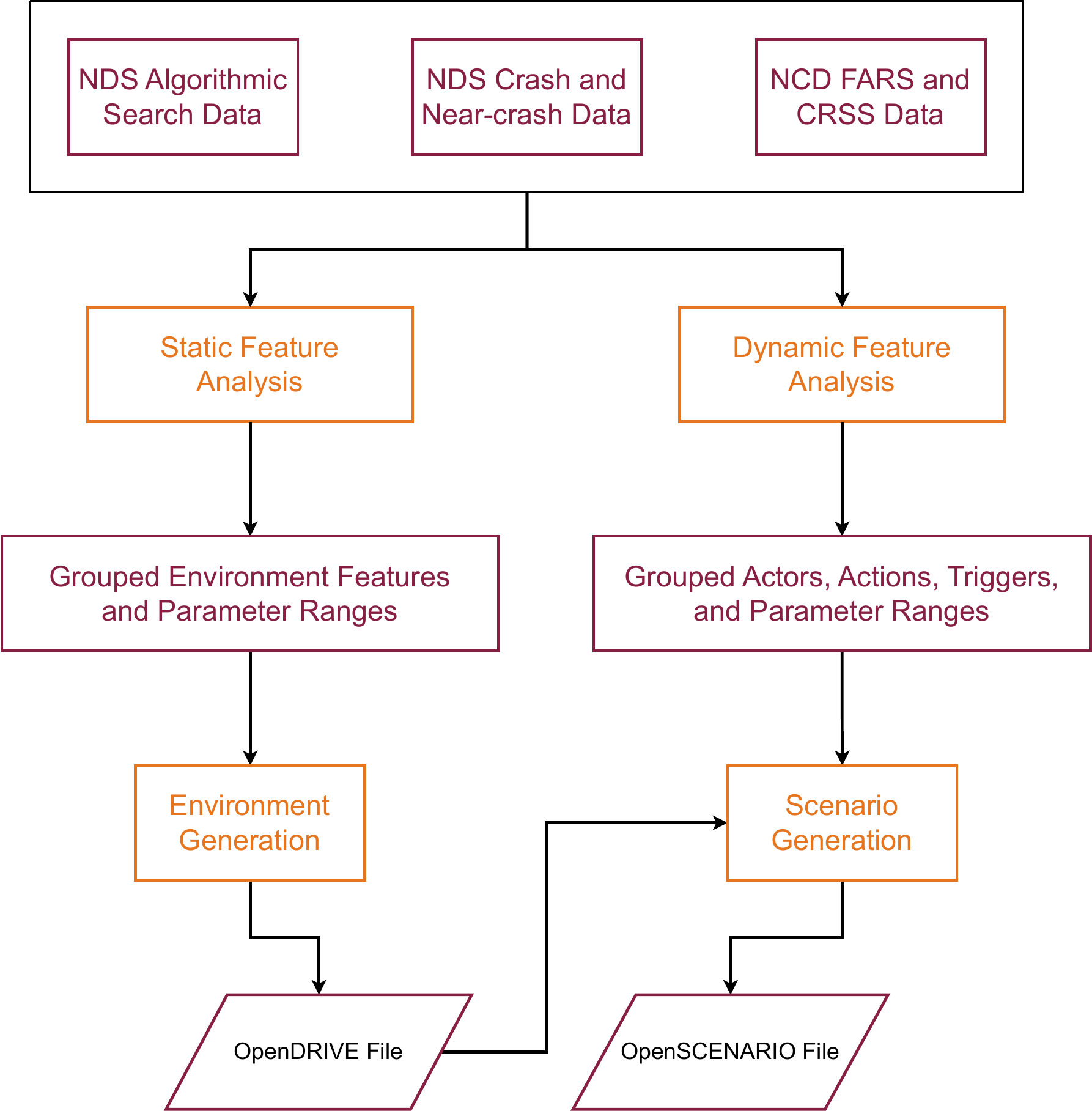}
\caption{Concrete test scenario generation process.}
\label{fig:scenario-gen}
\end{figure}

Figure \ref{fig:scenario-gen} illustrates the data-driven scenario generation process. The first step to concrete scenario generation is abstracting the important parameters required for inclusion when testing the system. The NCD and NDS data fusion performed in the previous steps provides the data to feed into this process. This step involves two processes; a static feature analysis and a dynamic feature analysis. The static feature analysis process groups the main roadway and environmental features such as number of lanes at the intersection, intersection control, and so forth. The dynamic feature analysis process groups actors, their initial states, their actions, and triggering conditions.

Once the groupings are created, combinations of interest are selected, and the parameter distributions are quantified. Representing the scenario this way can be effective in multiple ways. First, it helps normalize the data across different sets through parameter definitions and mappings. Second, it helps provide an understanding of the situational differences between safety-critical and non-critical events. Finally, this process provides a multi-dimensional parameter space that is reflective of how the scenario occurs in the real world. The analysis of the parameter space drives sampling strategies that allow for development of the test cases for use with simulation or for physical testing. Therefore, OpenDRIVE and OpenSCENARIO files representing an actual or hypothetical case representing a combination of parameters can be generated using this process.




\section{Discussion and Conclusions}

Scenario-based methodologies play a vital role in development and evaluation of ADAS/ADS functions. However, most current scenario-based methodologies have limitations in terms of representativity, coverage, and data richness.
Datasets used for scenario development often cover only part of the spectrum in which certain scenarios occur, and offer a limited number of variables to define the operational design domain (ODD), resulting in narrow and ambiguous  data representativeness. Understanding the range of data sources, from naturalistic driving data to multi-year record based national crash datasets is crucial for ensuring scenario representativeness. 

Our work addresses these challenges by offering representativity through data volume (spanning multiple years and millions of miles), richness in context (incorporating numerous record-based coded variables and augmented time series data), and severity coverage (encompassing routine driving to rare crash events). We employed a data-driven scenario-based approach, integrating human benchmarking strategies from crash and naturalistic driving data to support ADAS/ADS development and safety assessment. We introduced a comprehensive framework and associated tools utilizing a wealth of U.S.-based data sources, including 10 years of National Crash Data (NCD), over 34 million miles of naturalistic driving study (NDS) data, and various national- and state-level driving mileage datasets. 

This work effectively integrated and operationalized data from multiple sources while providing the context to show the complexity of the real-world driving environments. Since different datasets contain varying levels of information, our fused datasets encompass the full range of event severity (from routine driving to fatal crashes). They also contain both record-based variables as well as time series kinematic data which enables a more holistic  representation of the scenario. Additionally, data at each severity level provides valuable insights; crashes offer insights into failure modes, while routine driving events provide parameter distributions, comfort zones, and acceptance criteria.

As shown by our example analysis of ``turns at intersections'', our scenario fusion provides detailed insights into scenario frequencies, key parameter ranges, and relevant factors spanning driver behavior, vehicle characteristics, and environmental conditions. For example, the provided distributions of kinematic variables during turns (e.g., yaw and maxima/maximum lateral acceleration) can be used to seed scenario development in record-based datasets where such information is missing, develop multivariate distributions and models, select outlier and edge cases, and feed into clustering processes for test case generation. This information can also help system developers as well as safety experts to understand the severity versus frequency aspects of a scenario.

In conclusion, the integrated scenario-based analysis provides a holistic, data-driven, and representative approach to aid development and safety evaluation of ADAS/ADS. By curating and fusing various data sources, applying domain expertise to define scenarios, extracting key parameters, determining frequencies, and generating test cases, this approach ensures a thorough representation of real-world driving conditions. Future work will focus on refining this methodology by incorporating additional data sources and expanding the range of scenarios considered, thereby enhancing its robustness and applicability.

\section*{Acknowledgement}
The work presented in this paper has been funded by the Automated Mobility Partnership (AMP). AMP is a multi-year industry partnership promoting the development of tools, techniques, and data resources to support the advancement of vehicle technology for its members. The AMP program provides its members access to the curated NDS and NCD data presented in this paper, such that data interrogation can be performed, as well as data analytical tools to understand the important characteristics of the data at a case, scenario, or dataset level. The authors sincerely acknowledge all past and current AMP members for their helpful insights, feedback, discussions, and support.

\bibliography{ITSC_References.bib}
\bibliographystyle{ieeetr}

\vspace{12pt}

\end{document}